\documentclass[12pt]{bmvc2k}

\usepackage{booktabs}
\usepackage{multirow}
\usepackage{threeparttable}
\usepackage{float}
\usepackage{footnote}
\usepackage{amssymb}
\usepackage{geometry}
\title{Deep Scene Text Detection with Connected Component Proposals}

\addauthor{Fan Jiang*}{frank.jf@alibaba-inc.com}{1}
\addauthor{Zhihui Hao*}{hzh106945@alibaba-inc.com}{1}
\addauthor{Xinran Liu}{tom.lxr@alibaba-inc.com}{1}

\addinstitution{
 Amap Vision Lab,\\ 
 Alibaba Group,\\
 Beijing, China \\
*Contributed equally
}
\runninghead{AMAP VISION LAB}{Fan Jiang, Zhihui Hao, Xinran Liu}

\def\etal{\emph{et al}\bmvaOneDot}

\begin{document}

\maketitle

\begin{abstract}

A growing demand for natural-scene text detection has been witnessed by the computer vision community since text information plays a significant role in scene understanding and image indexing. 
Deep neural networks are being used due to their strong capabilities of pixel-wise classification or word localization, similar to being used in common vision problems. 
In this paper, we present a novel two-task network with integrating bottom and top cues. 
The first task aims to predict a pixel-by-pixel labeling and based on which, word proposals are generated with a canonical connected component analysis. 
The second task aims to output a bundle of character candidates used later to verify the word proposals.
The two sub-networks share base convolutional features and moreover, we present a new loss to strengthen the interaction between them.
We evaluate the proposed network on public benchmark datasets and show it can detect arbitrary-orientation scene text with a finer output boundary. 
In ICDAR 2013 text localization task, we achieve the state-of-the-art performance with an F-score of 0.919 and a much better recall of 0.915.

\end{abstract}

%-------------------------------------------------------------------------
% Introduction
%-------------------------------------------------------------------------
\section{Introduction}
\label{sec:intro}

Text information is an important supplementary clue for image understanding.
For example, if the words on a tag can be correctly read, the object in the image can often be identified and categorized immediately.
In natural scenes, text information is more meaningful for being an indicator of the observer's position. 
By reading text on shops and street signs, a scene image can be quickly aligned to digital maps and meanwhile, it gives the observer a fast and relatively precise localization.
Due to the increasing demands, text detection and reading have gained considerable attention over the past decade.

Deep convolutional networks, as in common vision problems, have achieved substantial success in text extraction as well.
In this paper, we focus on text detection in the wild with deep models.
As the first as well as a crucial step, scene text detection is challenging due to the cluttered backgrounds, uncontrolled illumination and strong variations in the text font and orientation.
Previous studies can be generally divided into two groups. 
Methods in the first group aim to obtain a pixel-wise labeling with the text as the foreground.
For example, fully convolutional networks can be applied to detect arbitrarily oriented text lines with the state-of-the-art performance \cite{zhang2016multi}.
In the second group, methods follow the object-detection paradigm and predict word bounding boxes, for example, DeepText \cite{zhong2016deeptext} and TextBoxes \cite{liao2016textboxes}.

\begin{figure}
\centering
\begin{tabular}{ccc}
\bmvaHangBox{\fbox{\includegraphics[width=3cm]{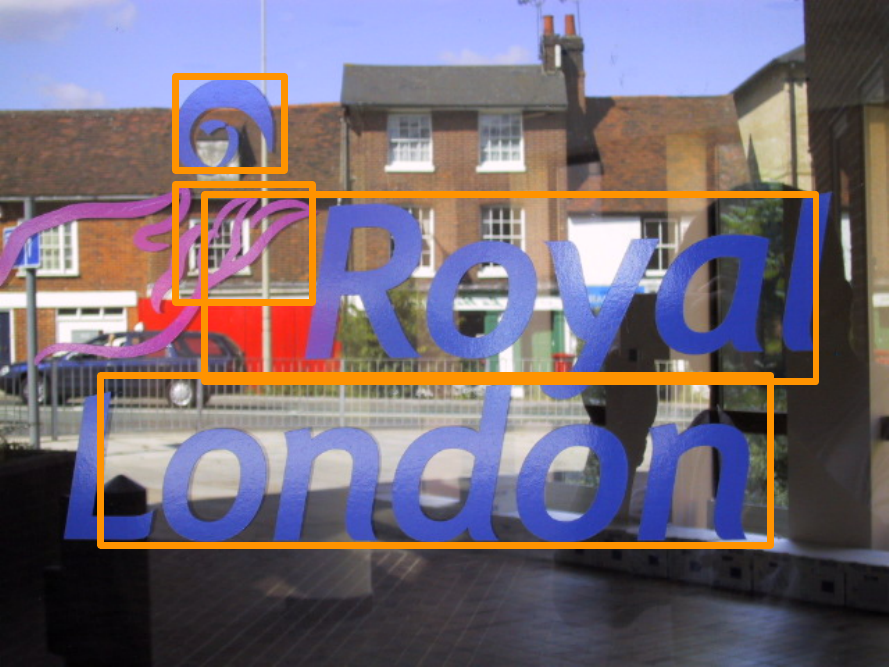}}}&
\bmvaHangBox{\fbox{\includegraphics[width=3cm]{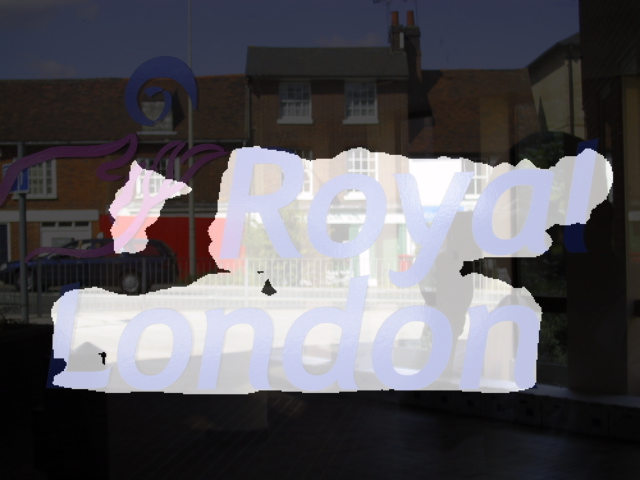}}}&
\bmvaHangBox{\fbox{\includegraphics[width=3cm]{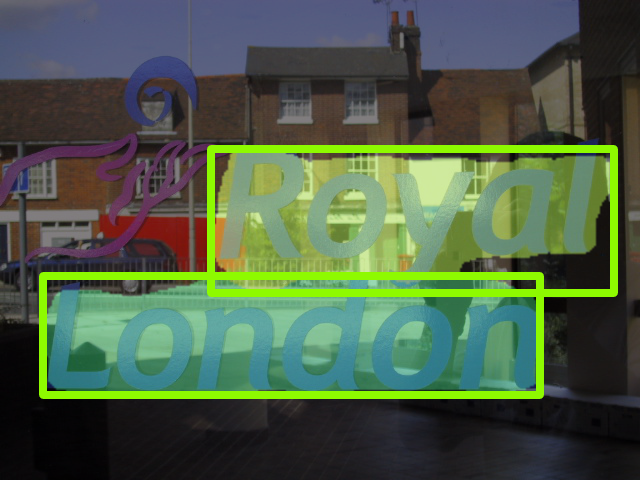}}}\\
(a)&(b)&(c)
\end{tabular}
\caption{
An example of previous methods and ours. 
Sub-figures show (a) the word detection by Faster R-CNN \cite{ren2015faster},
(b) the text segmentation by ParseNet \cite{liu2015parsenet},
and (c) the final prediction by our network.
}
\label{fig:badcases}
\end{figure}

As illustrated in Fig. \ref{fig:badcases}, we start with our observations that either paradigm has its own drawbacks.
When training an image parsing model, text lines or words may be labeled as quadrilaterals, but pixels in the holes of characters or between adjacent characters have no difference with the backgrounds.
This could be confusing for the model trainer.
Without any supervision from the structure information of characters, it may lead to a vague prediction on the contours of text and a poor measurement of precision.
However, when applying a Faster R-CNN \cite{ren2015faster} like detector, we also observe that the performance could be greatly hampered by the orientation variation and perspective distortion.

This paper focuses on the integration of the two types of deep models.
One contribution is that we experimentally show by sharing the convolutional features and training jointly, 
the proposed network achieves a better performance on scene text detection and instance separation.
Simply mixing the two tasks are not new \cite{brahmbhatt2016stuffnet, shrivastava2016contextual}, 
but for the first time, we confirm the benefits in text detection.
The second contribution is that we use the canonical connected component analysis (CCA) as a middle layer for proposal generation, \emph{before} making the final decision.
We present a new loss to explicitly enforce the consistency between the raw outputs of CCA layer and the characters detected.
Clearly, these efforts reduce false alarms and improve the text boundaries.

%\clearpage

%-------------------------------------------------------------------------
% Related Work
%-------------------------------------------------------------------------
\section{Related work}

A typical text reading pipeline, taking PhotoOCR \cite{bissacco2013photoocr} as an example, contains at least three modules: text detection, character recognition and word assembling.
Each of them can be implemented with several different methods.
Sequential learning networks, represented by Long Short-Term Memory \cite{hochreiter1997long} and Connectionist Temporal Classification \cite{graves2006connectionist} loss, recently show promising results in recognizing the text-line as a whole,
which can efficiently replace the latter two modules of the conventional pipeline.
In this sense, the performance of text detection becomes significant in terms of both recall and precision.

In this paper, we focus on the use of convolutional neural networks (CNNs) in scene-text detection.
It can date back to 2012, 
when Wang \etal \cite{wang2012end} presented a sliding-window approach to detect individual characters.
The convolutional network was being used as a 62-category classifier.
With the emergence of dedicated networks for common object detection,
applying those models into text problem seems straightforward.
In DeepText of Zhong \etal \cite{zhong2016deeptext}, they follow the Faster R-CNN \cite{ren2015faster} to detect words in images.
The Region Proposal Network is redesigned with the introduction of multiple sets of convolution and pooling layers.
The work of \cite{liao2016textboxes} follows another recent network called SSD \cite{liu2016ssd} with implicit proposals.
The authors also improve the adaption of the model to text issue by adjusting the network parameters.
The major challenge for word detection networks is the great variation of words in aspect ratio and orientation,
both of which can significantly reduce the efficiency of word proposals.
In the work of Shi \etal \cite{shi2016robust}, a Spatial Transformer Network \cite{jaderberg2015spatial} is introduced.
By projecting selected landmark points, the problem of rotation and  perspective distortion can be partly solved.

Another group of methods are based on image segmentation networks.
Zhang \etal \cite{zhang2016multi} use the Fully Convolutional Network (FCN) \cite{long2015fully} to obtain salient maps 
with the foreground as candidates of text lines.
The trouble is that the candidates may stick to each other,
and their boundaries are often blurry.
To make the final predictions of quadrilateral shape,
the authors have to set up some hard constrains regrading intensity and geometry.

%------------------------------------------------------------------------
% The Proposed Network
%------------------------------------------------------------------------
\section{The proposed network}

Our primary motivation of this work is to improve the performances of both text segmentation and character detection through joint training. 
First, we describe the architecture of the hybrid convolutional network in detail.
Then we present the key step of the proposal generating through the connected component analysis.
With the aim of suppressing the false alarms and enhancing the consistency, we show how the two tasks affect each other under the new loss.
The final outputs of the network are the connected components that have been verified by the character candidates.
Based on these segments, bounding boxes of words can be obtained with no effort.

%------------------------------------------------------------------------
% Network Architecture
%------------------------------------------------------------------------
\begin{figure}
\centering
\includegraphics[width=10.7cm]{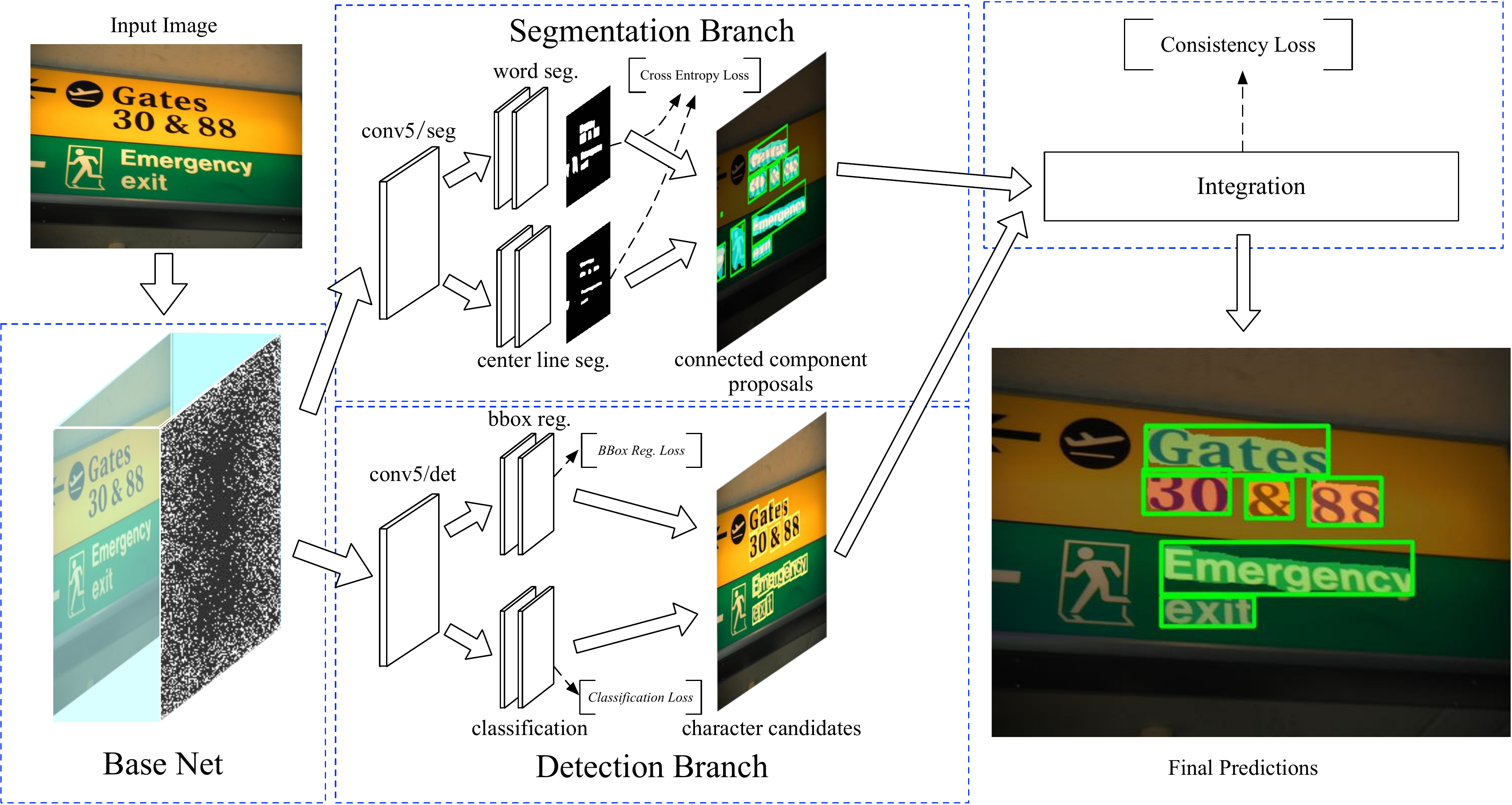}
\caption{
The architecture of the proposed network.
}
\label{fig:pipeline}
\end{figure}

\subsection{Network architecture}
\label{sub:architecture}

As showed in Fig. \ref{fig:pipeline}, the proposed deep network is composed of two branches: text segmentation and character detection.
They share the base convolution and pooling layers, which we implemented with VGG-16 \cite{simonyan2014very}.

The segmentation branch is similar to ParseNet \cite{liu2015parsenet}, which is trained to predict a pixel-wise text/non-text classification.
We add another sub-task into the branch with predicting the center line for each text region.
Both sub-tasks output saliency maps that ultimately indicate the probable presence of text instances.
These clues are collected by the subsequent connected component layer and are utilized to generate text proposals.

The detection branch is almost the same as the off-the-shelf Faster R-CNN \cite{ren2015faster} with the target of character detection.
It consists of the region-proposal part and a refinement part that predict simultaneously the category and the location of each character.
The detection task here is to produce a bunch of characters that are being used to verify the text proposals.
Put more simply, proposals where characters are found will pass through; otherwise, they will be suppressed as false alarms.

Note that training multiple tasks in one network is general and both branches can be implemented with several alternatives.
However, considering the relatively fixed aspect ratio, we argue that characters are more likely to be detected with a high recall than words.

%------------------------------------------------------------------------
% Connected Component Proposals and Layout Analysis
%------------------------------------------------------------------------
\subsection{Connected component proposals}

\begin{figure}
\centering
\includegraphics[width=10cm]{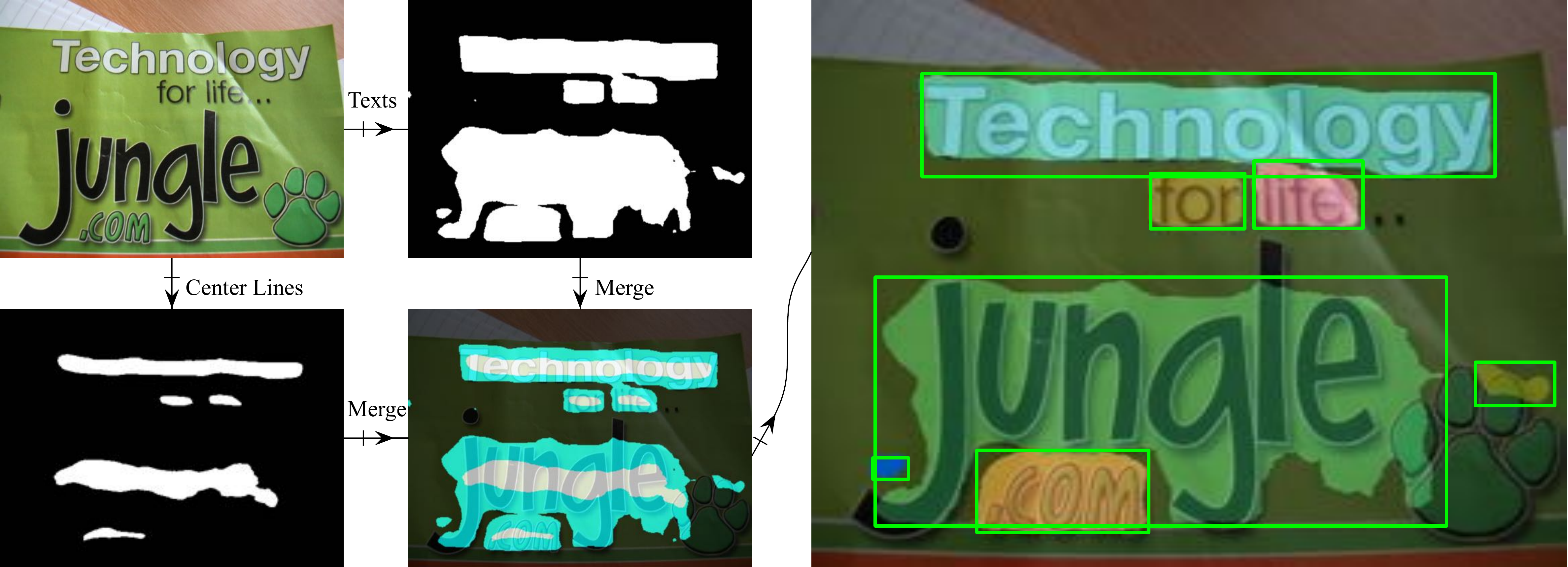}
\caption{
The generation process of connected component proposals.
}
\label{fig:cc-proposals}
\end{figure}

% para-1: center lines --> instance. 
The segmentation branch is trained with ground truth maps of \emph{words} and their center lines to predict them.
Details will be given in Sec. \ref{sec:experiment}.
We then use the output masks to produce proposals for word instances. 
The process is illustrated in Fig. \ref{fig:cc-proposals}.
Note that in a detection network, a proposal refers to a rectangle box generated as an object candidate to be confirmed.
Here a proposal means a group of pixels that connect each other and unlike character components \cite{yin2014robust}, each proposal here is a hypothesis of a whole word.

% para-2: pixels on the foreground, inner distance -->  the concept of proposals here  distance 
The prediction of center lines provides a favorable basis for word separation, since they are similar to the skeletons of words.
A connected component analysis is used to count these segments.
Then we take them as the cluster centers and divide all pixels on the word mask map into groups.
The criterion is based on the distance of a word pixel to its nearest center pixel.
By this means, the proposals of word instance are generated.

% para-3: segments --> bounding boxes
When calculating the distance of a path during the clustering of word pixels, shortest path problems are involved.
Our algorithm is similar to the Dijkstra's and the cost of traversing different segments is set as infinitely high.
In other words, pixels belonging to a connected component are preferred to be grouped together.
See the character \emph{j} in Fig. \ref{fig:cc-proposals}.

%------------------------------------------------------------------------
% Segmentation and Consistency Loss
%------------------------------------------------------------------------
\subsection{The consistency loss}

Now the network has two sets of proposals: word segments and characters.
They have to come to an agreement before the final decision is made.
Although it has been confirmed that by simply sharing the convolutional features, 
performances of different tasks can be mutually improved \cite{brahmbhatt2016stuffnet, shrivastava2016contextual}, we design a new loss to further strengthen their interactions.

The first part of the loss aims to penalize the false predictions of pixel in segmentation.
Pixels that are supported by character detections but are missing in the word saliency map cause a loss.
See Eq. \eqref{eq:parse_harmony_loss}, where $s_i$ is the binary value on the mask map of the word segmentation.
$\Phi_{det}$ defines the set of all pixels inside character boxes.
To reduce the influence of false detections, only the boxes with a high enough confidence are considered. 
Note that a word is naturally bigger than the sum of its characters, pixels outside all character boxes are hard to judge.
However, when a word segment has no character on it, all the pixels will be punished.
The set of these pixels is denoted as $\Psi_{seg}$.

\begin{equation} 
L_1 = \frac{1}{2N} \sum_{i}^{N} (1 - s_i) + \frac{1}{2M} \sum_j^M s_j, \ \ \ i \in \Phi_{det}, \ j \in \Psi_{seg}
\label{eq:parse_harmony_loss}
\end{equation} 

The second part of our consistency loss is designed to penalize the \emph{potential} false alarms in character detection.
As in Eq. \eqref{eq:proposal_harmony_loss}, $b_k$ indexes all the character boxes.
The function of $ratio(b_k)$ calculates for each box how much it is filled by the segmentation mask.
$\mathbb{I}(condition)$ denotes a boolean function.
When the ratio is lower than the threshold $\tau$, the box is not supported by pixel proposals and causes a loss.
Note that the real false alarms have been penalized in the detection sub-network, the loss here creates a chance to reinforce the punishments based on a pixel-level inspection.

\begin{equation} 
L_2 = \frac{1}{2K} \sum_k^K \mathbb{I}(ratio(b_k) < \tau)
\label{eq:proposal_harmony_loss}
\end{equation} 

\begin{figure}
\centering
\begin{tabular}{ccc}
\bmvaHangBox{\fbox{\includegraphics[width=3cm]{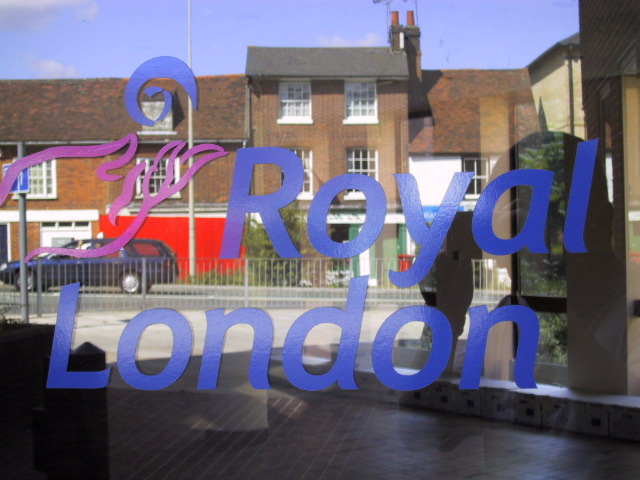}}}&
\bmvaHangBox{\fbox{\includegraphics[width=3cm]{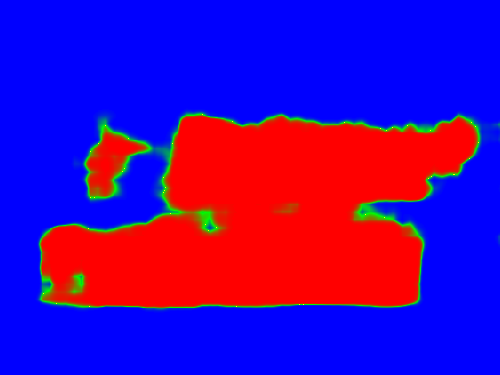}}}&
\bmvaHangBox{\fbox{\includegraphics[width=3cm]{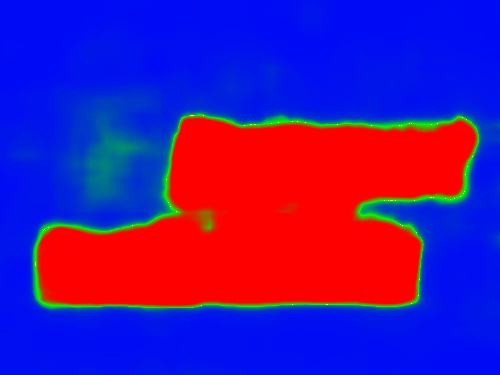}}}\\
(a)&(b)&(c)
\end{tabular}
\caption{The probability maps of segmentation trained (b) without and (c) with the consistency loss.}
\label{fig:harmony-loss}

\end{figure}

The consistency loss is a sum of the above two parts. 
It forces the two tasks to comply with each other.
Fig. \ref{fig:harmony-loss} shows an example of the segmentation results trained without and with this loss.
Same data are used and both trainings converge.
We can see under the consistency loss, a false alarm of segment is removed and the boundaries obviously become smoother.
To confirm the benefits, we quantitatively compare the network trained with this loss and its baseline without the loss.
See Tab. \ref{tab:performance_comparison} for details.

%------------------------------------------------------------------------
% Experiments
%------------------------------------------------------------------------
\section{Experiments}
\label{sec:experiment}

In this section, we first elaborate the network implementation and describe the two-staged training procedure.
Then we evaluate the performance of the proposed network on two datasets, namely the ICDAR 2013 \cite{karatzas2013icdar} Robust Reading competition and the Multilingual \cite{pan2011hybrid}. 
We have achieved the highest F-score on both datasets, compared with the previous state-of-the-arts.
We show some results on challenging images from these two datasets and Google Street View Text (SVT) \cite{wang2012end}.

\begin{table}[tp]
    \centering
    \fontsize{7}{9}\selectfont
    \begin{threeparttable}
    \caption{
    Results on the ICDAR 2013 Robust Reading Competition scene text dataset and the Multilingual dataset.
    The {\em Proposed} network is at the bottom line.
    The  {\em Baseline} is a counterpart trained almost the same but without  the consistency loss.
    }
    \label{tab:performance_comparison}
        \begin{tabular}{lccclccc}
        \toprule
        \multicolumn{1}{c}{\multirow{2}{*}{Method}}&
        \multicolumn{3}{c}{ICDAR 2013}&
%:
        \multicolumn{1}{c}{\multirow{2}{*}{Method}}&
        \multicolumn{3}{c}{Multilingual}\cr
        \cmidrule(lr){2-4} \cmidrule(lr){6-8}
        &Recall&Precision&F-score&&Recall&Precision&F-score\cr
        \midrule
        
        TextBox-v2 \cite{liao2016textboxes}&	0.867&	0.919	&	0.886&		Pan \cite{pan2011hybrid}			&	0.659	&	0.645	&	0.652\cr
        CTPN \cite{tian2016detecting} 		&	0.830&	0.930	&	0.877&		Yin \cite{yin2014robust}			&	0.685	&	0.826	&	0.749\cr
        DeepText \cite{zhong2016deeptext}&	0.842&	0.907	&	0.873&		StradVison \cite{zhou2015icdar}	&	0.719	&	0.815	&	0.764\cr
        Zhu \cite{zhu2016text} 			&	0.816&{\bf0.934}	&	0.871&		TextFlow \cite{tian2015text}		&	0.784	&	0.847	& 	0.814\cr
        Qin \cite{qin2017cascaded}		&	0.824&	0.891	&	0.856&		CTPN \cite{tian2016detecting}		&	0.800 	&{\bf0.840}	& 	0.820\cr
        StradVison \cite{zhou2015icdar}	&	0.802&	0.909	&	0.852&		&	 	&	 	&\cr
        \hline
        Baseline						&	0.903&	0.913	&	0.908&		Baseline						&	0.836	&	0.812	&	0.824\cr
        the Proposed 					&{\bf0.915}&	0.922	&	 {\bf0.919}&	the Proposed					&{\bf 0.843}	&	0.809	&{\bf0.826}\cr
        \bottomrule
        \end{tabular}
    \end{threeparttable}
\end{table}

\subsection{Implementation and training details} 

The network is a hybrid one with ParseNet and Faster R-CNN as shown in Fig. \ref{fig:pipeline}.
Following the ParseNet and Faster R-CNN implemented with the Caffe framework \cite{jia2014caffe}, we modify the parameters and architectures to build the proposed network.
The base convolution layers from {\tt conv1\_1} to {\tt conv4\_3} are shared by ParseNet and Faster R-CNN.
In consideration of relatively different targets of segmentation and detection, the two sub-networks hold individual convolution layers from {\tt conv5\_1} to {\tt conv5\_3} to build better generalization capability. 
We add a sub-task in ParseNet to predict the center lines of text regions. 
The sub-task shares the convolution layers from {\tt fc6/seg} to {\tt fc7/seg} with the pixel-wise text/non-text classification task and duplicate the subsequent layers of the ParseNet.

To train the proposed network, we prepare the training set with three types of ground-truth, which are the text regions, the center lines and the characters. 
The text regions and center lines are used to train the double-task ParseNet.
The text regions are labelled in word-level with polygons.
The center lines are labelled as 50\% height of the text regions.
The characters are labelled with their bounding rectangles and 95 classes, including background and 94 ASCII characters. 
The characters are then used to train the character detection Faster R-CNN. 

We perform a two-staged training to gain the final model. On the first stage, we parallelly train the ParseNet and Faster R-CNN following the conductions from the original works. 
Since the final proposals are made by the ParseNet sub-network, we stitch the subsequent layers of {\tt conv4\_3} from the Faster R-CNN to the ParseNet. 
While training the ParseNet, we apply a cross entropy loss to simultaneously control the two tasks of  text regions and center lines predicting.

On the second stage, we train with a momentum of 0.9, weight decay of 0.0005. 
We first set the {\em lr\_mult} of the base convolution layers and ParseNet layers to $0$ and train the Faster R-CNN branch with $10^{-6}$ learning rate on 30K SGD iterations.
Then we reset the {\em lr\_mult} and set the learning rate to $10^{-8}$ on 10K SGD iterations and followed by another 30K SGD iterations with $10^{-9}$ learning rate.

\subsection{Evaluation and discussion}

\begin{figure}
\centering
\includegraphics[width=11cm]{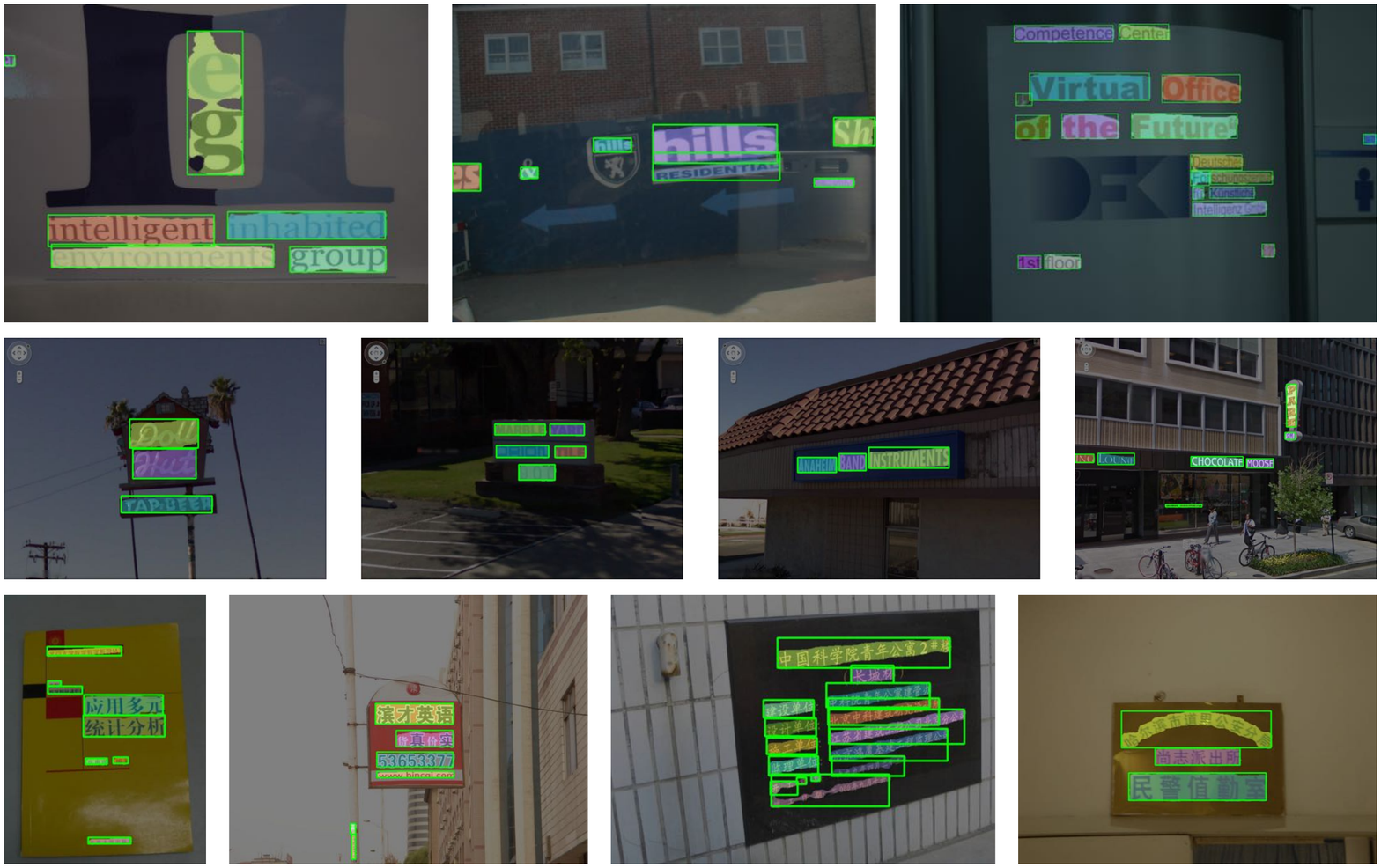}
\caption{Examples of scene text detection with our network. Images of each row are from the ICDAR 2013 Focused Text, the SVT and the Multilingual datasets.}
\label{fig:other-results}
\end{figure}

For the evaluation, we follow the standard protocols provided by the dataset creators or competition organizers. 

\noindent{\bf ICDAR 2013 dataset.}\ \ 
The ICDAR 2013 text localization task consists of images with text as the main objects, namely "Focused Text" \cite{karatzas2013icdar}.
The dataset provides 229 training images and 233 testing images. 
The images include text of English characters, digits and punctuations. 
We follow the online evaluation system 
\footnote{Available on http://rrc.cvc.uab.es/?ch=2\&com=introduction}
under the protocol of "Deteval", which is provided by the competition organizer. 
See Tab. \ref{tab:performance_comparison}, the proposed network achieves the rank-1st performance with a highest F-score and a much better recall.

\noindent{\bf Multilingual dataset.}\ \ 
The Multilingual dataset is presented in \cite{pan2011hybrid} with 248 training images and 239 testing images. 
The images are labelled in text line level with a mixed typesetting of Chinese, English and digits. 
In this dataset we revise our labeling strategy to text line level and append a class, namely Chinese, to the character detection branch.
The performance behaves similarly as ICDAR 2013, besides the high F-score, the proposed method acts with a noticeable high recall.

\noindent{\bf Discussion.}\ \ 
Note that the candidates are generated by pixel-wise classification, we argue that the proposed network offers a performance with a high recall due to the merit of segmentation-based proposals, which are able to extract text in any shape and arbitrary orientation.
This is also confirmed on non-latin texts.
Comparing the performance of the {\em Proposed} network with the {\em Baseline}, we have the reasonable ground to believe that  with the consistency loss,
the performance is further improved.
We also believe that if replace the VGG-16 with a deeper network as the base layers, the proposed network can achieve an even higher score.

%------------------------------------------------------------------------
% Conclusions
%------------------------------------------------------------------------
\section{Conclusions}

In this paper, we propose a deep neural network for scene text detection.
The network consists of two branches, with the segmentation branch generating the text proposals with connected component analysis and the detection branch verifying these proposals with a new consistency loss.
Using the proposed network, we have obtained the best performance in the text localization task of ICDAR 2013. % Change one.
As in many other vision problems, we believe the integration of pixel-level bottom cues and object-level top cues is significant for text detection as well, and this paper provides a compelling evidence in this direction.

%-------------------------------------------------------------------------

\bibliography{egbib}

\end{document}